\documentclass[10pt,twocolumn,letterpaper]{article}

\usepackage{multirow}
\usepackage[pagenumbers]{cvpr} 
\usepackage[most]{tcolorbox}

\usepackage[T1]{fontenc}
\usepackage[utf8]{inputenc}
\usepackage{listings}
\definecolor{cvprblue}{rgb}{0.21,0.49,0.74}
\usepackage[pagebackref,breaklinks,colorlinks,allcolors=cvprblue]{hyperref}
\usepackage{tabularx}

\def\paperID{*****} 
\def\confName{CVPR}
\def\confYear{2026}

\title{LaScA: Language-Conditioned Scalable Modelling  of Affective Dynamics}

\author{Kosmas Pinitas\\
Department of Digital Systems\\
University of Piraeus\\
{\tt\small kpinitas@unipi.gr}
\and
Ilias Maglogiannis\\
Department of Digital Systems\\
University of Piraeus\\
{\tt\small imaglo@unipi.gr}
}

\begin{document}
\maketitle
\begin{abstract}
Predicting affect in unconstrained environments remains a fundamental challenge in human-centered AI. While deep neural embeddings dominate contemporary approaches, they often lack interpretability and limit expert-driven refinement. We propose a novel framework that uses Language Models (LMs) as semantic context conditioners over handcrafted affect descriptors to model changes in Valence and Arousal. Our approach begins with interpretable facial geometry and acoustic features derived from structured domain knowledge. These features are transformed into symbolic natural-language descriptions encoding their affective implications. A pretrained LM processes these descriptions to generate semantic context embeddings that act as high-level priors over affective dynamics. Unlike end-to-end black-box pipelines, our framework preserves feature transparency while leveraging the contextual abstraction capabilities of LMs. We evaluate the proposed method on the Aff-Wild2 and SEWA datasets for affect change prediction. Experimental results show consistent improvements in accuracy for both Valence and Arousal compared to handcrafted-only and deep-embedding baselines. Our findings demonstrate that semantic conditioning enables interpretable affect modelling without sacrificing predictive performance, offering a transparent and computationally efficient alternative to fully end-to-end architectures.

\end{abstract}    
\section{Introduction}

Modelling affective behaviour in unconstrained, in-the-wild environments remains a central challenge in affective computing \cite{zafeiriou2016facial,makantasis2023lab}. Continuous estimation of valence and arousal provides a compact representation of emotional state; however, annotations collected in naturalistic settings are inherently noisy and subjective \cite{kollias2021affect, barthet2023knowing}. Inter-annotator disagreement, cultural variability, and contextual ambiguity make absolute affect values difficult to estimate reliably, particularly when modelling affective dynamics over time. To mitigate annotation noise, affect modelling is often reformulated in directional terms, predicting whether affect increases or decreases between consecutive temporal segments \cite{lotfian2016practical}. This relative formulation emphasises temporal trends rather than precise magnitudes and has become a structured protocol for analysing affective trajectories in large-scale datasets \cite{pinitas2022rankneat,pinitas2024varying}.

Despite this structured formulation, constructing robust affect models in real-world conditions remains challenging. State-of-the-art approaches typically rely on end-to-end deep architectures that learn high-dimensional latent representations directly from visual and acoustic streams \cite{rouast2019deep}. While effective, these models often entangle signal extraction and affect reasoning within opaque embeddings, making it difficult to analyse how specific behavioural cues contribute to predictions and limiting the integration of structured domain knowledge \cite{yang2023survey}. In contrast, handcrafted facial geometry and acoustic descriptors provide compact and computationally efficient representations grounded in expert knowledge. However, these features lack contextual abstraction: the affective meaning of a facial movement or vocal modulation often depends on surrounding behavioural cues and temporal context. As a result, purely feature-based models struggle to capture the higher-level semantic relationships underlying affective dynamics \cite{mishra2015facial}.

To address this limitation, we introduce \textit{LaScA}, a framework that integrates semantic knowledge from large language models into lightweight affect prediction pipelines. Rather than replacing structured behavioural features with opaque embeddings, LaScA preserves handcrafted facial and acoustic descriptors and augments them through language-conditioned representations. For each feature, a short linguistic description capturing its behavioural and affective implications is generated offline using a pretrained \emph{Large Language Model}. These descriptions are created once and stored as a deterministic semantic lexicon.

For each temporal segment, the descriptions corresponding to salient features are assembled into structured templates and encoded using a pretrained sentence transformer to produce semantic embeddings capturing contextual relationships among behavioural cues. These embeddings are fused with the original handcrafted features and used to model directional changes in valence and arousal through a lightweight preference learning module. Importantly, all language and representation encoders remain frozen, meaning that only a small prediction head is trained. This design keeps the model compact and enables efficient and scalable learning.

We evaluate LaScA on the Aff-Wild2 \cite{kollias2018aff} and SEWA \cite{kossaifi2019sewa} datasets under the affect change prediction protocol. Experiments demonstrate that language-conditioned representations consistently improve affect change prediction across visual, audio, and multimodal settings while remaining competitive with substantially larger deep embedding models. Ablation studies further show that these gains arise from structured language conditioning rather than increased model capacity. Overall, LaScA provides an efficient and reproducible approach for modelling affective dynamics in real-world environments.
\section{Related Work}

\subsection{Affective Dynamics and Multimodal Modelling}

Affect modelling has traditionally relied on handcrafted facial and acoustic descriptors, such as facial action units, head pose, and prosodic features, combined with classical machine learning models to estimate emotional states either as discrete categories or continuous valence–arousal signals~\cite{dhall2018aff, valstar2016avec,pinitas2022supervised}. While these descriptors are computationally efficient and grounded in domain knowledge, they often struggle to capture the complex contextual dependencies required for robust affect modelling in real-world conditions.

With the advent of deep learning, end-to-end architectures such as convolutional neural networks (CNNs), recurrent neural networks (RNNs), and Transformer-based models have become the dominant paradigm for affect prediction, learning latent representations directly from visual and auditory streams~\cite{, kollias2019deep, yousefi2025deeparousal}. Multimodal approaches that combine video, audio, and textual information further improve performance by exploiting complementary cues across modalities~\cite{kollias2022abaw, sharma2020survey, ahire2025maven, pinitas2025privileged, makantasis2022invariant}. However, these models often rely on large latent representations that make it difficult to analyse how individual behavioural cues influence predicted affective states.

To reduce sensitivity to annotation noise and better capture temporal evolution, several studies reformulate affect modelling as a directional prediction task, estimating whether affect increases or decreases between temporal segments~\cite{pinitas2022rankneat, naini2025ranklist, cho2025predicting}. Building on this formulation, our work combines structured behavioural features with language-conditioned semantic representations, enabling a lightweight and scalable framework for modelling affective dynamics.

\subsection{Language-Grounded Representations for Affect Modelling}

Recent work has explored how structured semantic knowledge can complement behavioural representations in multimodal learning. In affective computing, earlier research investigated symbolic reasoning, rule-based systems, and structured feature representations to analyse how behavioural cues relate to emotional states~\cite{yuvaraj2025affective, zhang2025exploring, wan2025survey}. Although facial and acoustic descriptors remain compact and grounded in domain knowledge, they often lack the capacity to capture higher-level contextual relationships that influence affective perception.

Large language models (LLMs) provide a promising mechanism for incorporating semantic knowledge into multimodal modelling pipelines. Recent works rely on pretrained language models to contextualise visual and auditory signals, often acting as cross-modal semantic priors or shared representation spaces~\cite{radford2021learning, zhang2023video, zeng2025analyst}. These approaches demonstrate that language representations can help align heterogeneous modalities and capture richer contextual relationships. However, many such methods integrate language models directly into end-to-end architectures or rely on prompt-based inference during training, which can increase computational complexity and introduce stochastic behaviour.

In this work, we explore a complementary strategy in which language models provide deterministic semantic grounding for structured affect descriptors. Specifically, we generate concise textual descriptions for each behavioural feature using a pretrained LLM and store them as a fixed semantic lexicon. These descriptions are then composed into per-sample textual representations and encoded using pretrained sentence transformers. By combining structured behavioural features with language-conditioned representations, our approach enables a lightweight and scalable framework for modelling affective dynamics while preserving the structure of the original descriptors.
\section{Methodology}

\begin{figure*}[t]
\centering
\includegraphics[width=0.83\textwidth]{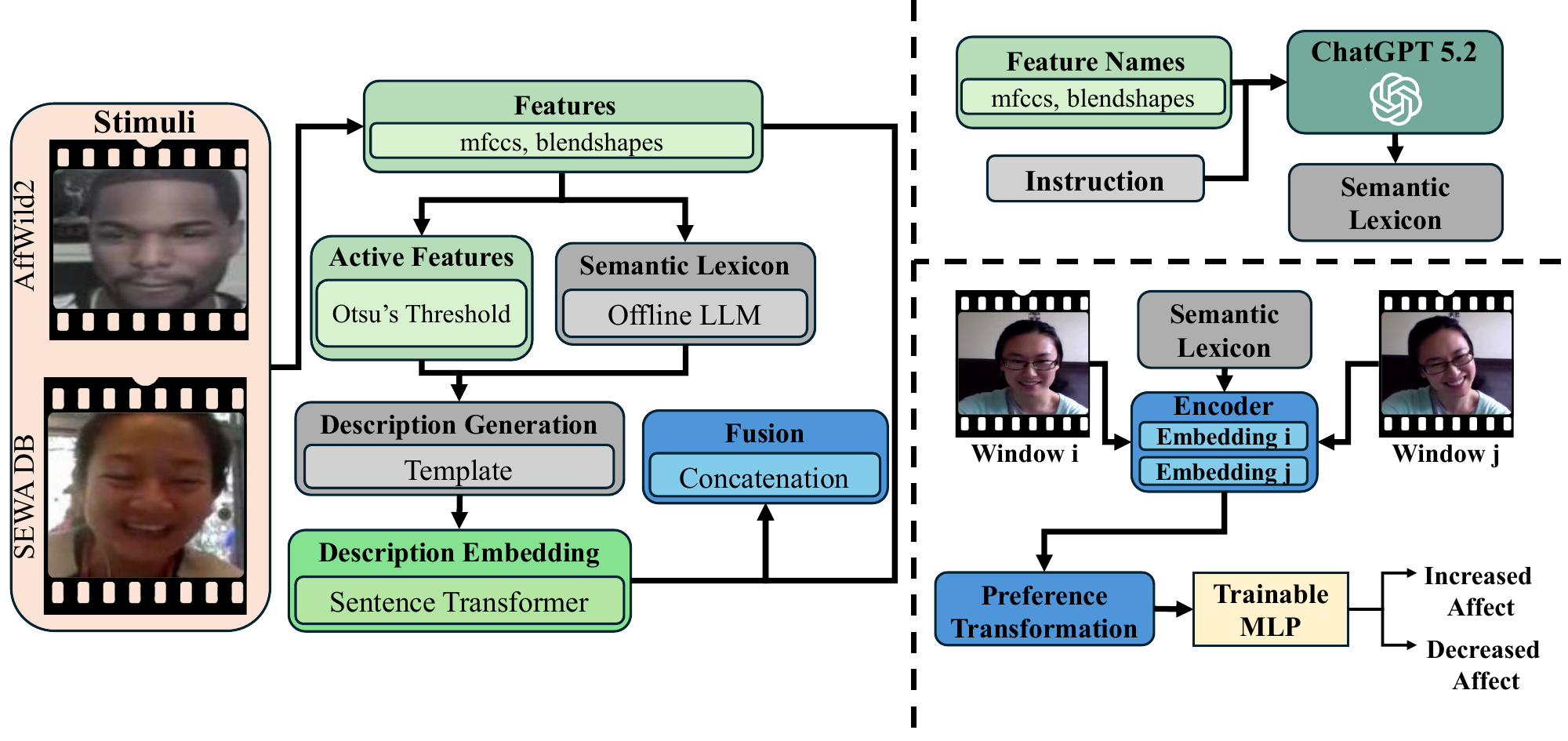}
\caption{
Overview of the LaScA framework. 
Handcrafted facial and acoustic features are first extracted and filtered via data-driven salience estimation (Otsu’s threshold). Each feature is associated with a fixed affect-aware semantic description generated offline by a large language model, forming a deterministic semantic lexicon. Active feature descriptions are composed into structured templates and encoded using a frozen sentence transformer to produce semantic embeddings. These embeddings are fused with the original feature representations and used to construct preference pairs between temporal windows. A trainable MLP operates on the transformed pairwise representations to predict directional affect change (increase or decrease).
}
\label{fig:framework}
\end{figure*}

This preliminary study proposes a hybrid affect modelling framework that combines handcrafted features with fixed semantic priors derived from pretrained language models. By decoupling signal transparency from contextual abstraction, our approach integrates  feature activations with a deterministic LLM-based lexicon, providing language-informed guidance over affective dynamics.

\subsection{The LaScA Framework}

We propose a deterministic framework that augments transparent affect descriptors with language-based semantic conditioning. The method converts continuous behavioural measurements into a sparse set of salient cues, grounds them in an affect-aware linguistic space, and fuses the resulting semantic representation with the original features for downstream affect change modelling (Figure \ref{fig:framework}).

\paragraph{Feature Extraction and Normalisation:}
For each temporal segment $t$, we extract structured visual and acoustic descriptors forming a feature vector
\begin{equation}
\mathbf{x}_t \in \mathbb{R}^d .
\end{equation}
To ensure comparability across dimensions, each feature is min--max normalised to the range $[0,1]$ using statistics computed on the training set.

\paragraph{Per-Sample Salience Estimation:}
To identify dominant behavioural cues within a segment, we estimate a data-driven partition of features into dominant and non-dominant groups. Specifically, we rank the normalised feature values within $\mathbf{x}_t$ and apply Otsu’s threshold to obtain a split that maximises between-group variance. This yields a binary activation mask
\begin{equation}
\mathbf{m}_t \in \{0,1\}^d,
\end{equation}
where $m_{t,i}=1$ denotes that feature $f_i$ is considered salient for segment $t$.  
We interpret this procedure as an unsupervised estimate of relative behavioural salience, enabling adaptive cue selection under strong inter-subject variability.

\paragraph{Affect-Aware Semantic Lexicon:}
Each feature is associated with a short textual description capturing its affective implication. The mapping
\begin{equation}
\mathcal{L} = \{ (f_i, \ell_i) \}_{i=1}^{d}
\end{equation}
is constructed offline using ChatGPT~5.2 by prompting the model to act as an affective computing researcher and produce standardised, domain-consistent descriptions for all features (Appendix \ref{app:llm_prompt}).
This lexicon is generated once and remains fixed throughout training and evaluation, ensuring reproducibility and eliminating stochastic language model effects (Appendix \ref{app:audio_desc}, \ref{app:facail_desc}).

\paragraph{Semantic Encoding:}
For each segment, the descriptions corresponding to active features are inserted into a structured template to form a textual representation (Appendix \ref{app:description_template}).
\begin{equation}
T_t = \mathrm{Template}(\{\ell_i \mid m_{t,i}=1\}).
\end{equation}
The text is encoded using a pretrained sentence transformer to obtain a semantic embedding
\begin{equation}
\mathbf{s}_t = \mathrm{Enc}(T_t),
\end{equation}
which captures contextual relationships among the observed cues in a shared linguistic space.

\paragraph{Representation Fusion:}
The final segment representation is obtained by concatenating handcrafted and semantic features
\begin{equation}
\mathbf{z}_t = [\mathbf{x}_t \, \| \, \mathbf{s}_t ],
\end{equation}
which is subsequently used by the preference learner to estimate affective change.

\subsection{Representation Components}\label{sec:encoders}

In representation learning, an encoder is a neural network that transforms raw or structured inputs into compact, high-level representations capturing task-relevant information. In this work, we investigate the effectiveness of language-conditioned representations for modelling affect dynamics and compare them against conventional unimodal baselines.

The proposed language encoder is a pretrained sentence transformer that converts the constructed textual template for each segment into a fixed-dimensional embedding. The encoder remains frozen during training to preserve the semantic structure induced by the affect-aware lexicon and to prevent overfitting to dataset-specific correlations. The resulting embedding captures contextual relationships among the activated behavioural cues within a shared linguistic space, serving as a semantic prior for affect modelling.

\paragraph{Sentence Transformers:} We evaluate five pretrained Sentence-Transformer encoders \cite{liu2019roberta,sanh2019distilbert,reimers2019sentence,song2020mpnet}: \textit{all-mpnet-base-v2}, \textit{multi-qa-mpnet-base-dot-v1}, \textit{all-distilroberta-v1}, \textit{all-MiniLM-L12-v2}, and \textit{multi-qa-distilbert-cos-v1}. These models cover multiple backbone architectures (MPNet, RoBERTa, MiniLM, DistilBERT) and embedding sizes (384–768). All encoders are used as frozen feature extractors. Given an input sentence $x$, each model produces an embedding $\mathbf{z} \in \mathbb{R}^d$ via its pooling mechanism, followed by an identical lightweight prediction head to ensure fair comparison (Table \ref{tab:sentence_models}).

\begin{table}[t]
\centering
\caption{Pretrained sentence encoders evaluated in this work.}
\label{tab:sentence_models}
\footnotesize
\setlength{\tabcolsep}{3pt}
\renewcommand{\arraystretch}{1.05}
\begin{tabular}{l l c c}
\toprule
\textbf{Model} & \textbf{Backbone} & \textbf{Pooling} & \textbf{Dim} \\
\midrule
all-mpnet-base-v2            & MPNet         & Mean & 768 \\
multi-qa-mpnet-base-dot-v1   & QAMPNet         & CLS  & 768 \\
all-distilroberta-v1         & DistilRoBERTa & Mean & 768 \\
all-MiniLM-L12-v2            & MiniLM        & Mean & 384 \\
multi-qa-distilbert-cos-v1   & DistilBERT    & Mean & 768 \\
\bottomrule
\end{tabular}
\end{table}

\begin{table}[t]
\centering
\caption{Pretrained baseline encoders used for benchmarking the proposed framework.}
\label{tab:benchmark_encoders}
\footnotesize
\setlength{\tabcolsep}{3pt}
\renewcommand{\arraystretch}{0.9}
\begin{tabular}{l l l c}
\toprule
\textbf{Visual} & \textbf{Backbone} & \textbf{Pooling} & \textbf{Dim} \\
\midrule
VGGFace2  & CNN & Global Avg Pool & 2048 \\
SwinFace & Swin Transformer & Token Pooling & 768 \\
MAE-Face & Transformer (MAE) & CLS Token & 768 \\
\bottomrule
\textbf{Audio} & \textbf{Backbone} & \textbf{Pooling} & \textbf{Dim} \\
\midrule
Wav2Vec2 & Wav2Vec2 Base & Mean over frames & 768 \\
MAE-Audio & Transformer (MAE) & CLS / Global Pool & 512 \\
\bottomrule
\textbf{Multimodal} & \textbf{Backbone} & \textbf{Pooling} & \textbf{Dim} \\
\midrule
MMA-DFER & Multimodal Fusion & Cross-modal & 1024 \\
HiCMAE & Hierarchical MAE & Avg Pool / CLS & 768 \\
\bottomrule
\end{tabular}
\end{table}

\paragraph*{Benchmark Encoders:} We benchmark against three pretrained face representation models: VGGFace2 ResNet50~\cite{cao2018vggface2}, SwinFace~\cite{qin2023swinface}, and MAE-Face~\cite{ma2022facial}. These models span convolutional, transformer-based, and self-supervised masked autoencoding paradigms, respectively, and have demonstrated strong performance on face-related tasks. All encoders are used as frozen feature extractors, and their embeddings are passed through an identical downstream prediction head to ensure a controlled comparison focused on representational quality rather than task-specific fine-tuning. For the \textbf{audio modality}, we employ two strong self-supervised models: Wav2Vec2~\cite{baevski2020wav2vec} and MAE-Audio~\cite{huang2022amae}. Wav2Vec2 learns speech representations from raw waveforms via contrastive learning, while MAE-Audio applies masked autoencoding to spectrogram representations. Both backbones are frozen and followed by a shared pooling and prediction head, ensuring fair comparison across methods and modalities. Finally, we compare against two recent state-of-the-art \textbf{multimodal frameworks} MMA-DFER~\cite{chumachenko2024mma} and HiCMAE~\cite{sun2024hicmae}. These approaches learn joint audio-visual representations through cross-modal alignment and self-supervised objectives. For consistency, we use their released backbone representations and apply the same downstream prediction head. (Table \ref{tab:benchmark_encoders}).

\subsection{The Preference Learner}
\label{sec:preference_learner}

We adopt a preference learning (PL) formulation that models ordinal relationships between consecutive samples rather than predicting absolute affect scores. This is well-suited for affect modelling, where relative changes are often more reliable than small absolute annotation differences.

Given two consecutive time windows $(x_t, x_{t+1})$ with ground-truth affect values $a_t$ and $a_{t+1}$, we construct preference pairs only when their relative change exceeds a margin $\tau$:
\[
\frac{|a_{t+1} - a_t|}{\max(|a_t|, \epsilon)} > \tau.
\]
Pairs with small variations are discarded to reduce ambiguity. For valid pairs, we assign a binary label indicating whether affect increases:
\[
y_{t,t+1} =
\begin{cases}
1 & \text{if } a_{t+1} > a_t \\
0 & \text{otherwise}.
\end{cases}
\]

Each sample is mapped to a latent embedding $\mathbf{z}_t$ using a frozen pretrained encoder. We model directional change via the embedding difference $\Delta \mathbf{z}_{t,t+1} = \mathbf{z}_{t+1} - \mathbf{z}_t$, which is passed through a lightweight two-layer MLP with sigmoid activation to predict the probability of affect increase. The model is trained using binary cross-entropy over constructed preference pairs. The margin $\tau$ controls a trade-off between robustness and data efficiency: larger values reduce label noise but decrease the number of training pairs, while smaller values increase supervision at the risk of incorporating ambiguous comparisons. We evaluate two relative thresholds (10\% and 20\%) to analyse this trade-off empirically.

\section{Case Study: Predicting Affect In-The-Wild}

We evaluate the proposed framework on two large-scale in-the-wild affect datasets, Aff-Wild2~\cite{kollias2019affwild2} and SEWA DB~\cite{kossaifi2019sewa}. Both provide continuous valence and arousal annotations under unconstrained recording conditions, making them suitable benchmarks for assessing robustness in realistic scenarios.

\subsection{Datasets}

\paragraph{Aff-Wild2:} Aff-Wild2~\cite{kollias2019affwild2} is one of the largest publicly available in-the-wild affect datasets. It consists of 558 videos collected from YouTube, featuring 458 subjects recorded under highly unconstrained conditions. The dataset contains approximately 2.8 million annotated frames for continuous valence and arousal prediction. Videos exhibit substantial variability in head pose, illumination, occlusions, camera motion, background clutter, and spontaneous facial behaviour. Continuous valence and arousal annotations are provided at the frame level by three independent annotators per video. Final ground-truth signals are obtained by aggregating annotator traces and computing the mean value $\mu$ across raters at each time step. Aff-Wild2 provides synchronised visual (face video) and audio streams, enabling unimodal and multimodal affect modelling under realistic web-scale conditions.

\paragraph{SEWA DB:}
SEWA DB~\cite{kossaifi2019sewa} (Sentiment Analysis in the Wild) is a multicultural audio-visual corpus collected to study spontaneous affective behaviour in dyadic interactions. The dataset contains video-chat recordings of pairs of participants from multiple cultural backgrounds (British, German, Hungarian, and Chinese) engaging in structured advertisement discussion tasks. These interactions are designed to elicit natural emotional responses while preserving conversational realism. The corpus provides continuous annotations for valence and arousal. Each recording is annotated independently by six raters, and final affect signals are computed as the mean $\mu$ across annotators to mitigate individual bias and improve temporal stability. SEWA DB includes synchronised facial video, audio signals, and facial landmark tracks. Compared to Aff-Wild2’s large-scale web videos, SEWA DB offers semi-controlled conversational recordings with cross-cultural variability, providing a complementary evaluation setting for affect modelling.

\subsection{Preprocessing}

\textbf{Feature Extraction:} For the handcrafted modality setting, we extract 58 facial blendshape coefficients per frame using MediaPipe Face Mesh, providing a compact and pose-robust representation of facial muscle activations. From the audio stream, we compute 15 Mel-Frequency Cepstral Coefficients (MFCCs) per time step to capture short-term spectral characteristics relevant to affective speech.  Audio and visual features are temporally aligned with frame-level annotations and segmented into fixed-length windows (3s and 5s). Within each window, features are mean-pooled to obtain a single window-level representation. Continuous valence and arousal annotations are similarly averaged, producing window-level affect values $\{a_t\}$.

\paragraph{Pretrained Encoder Processing:}
For pretrained baselines (Section~\ref{sec:encoders}), we follow each model’s original preprocessing protocol while adapting temporal segmentation to our 3s/5s windowing scheme. Visual backbones operate on cropped and normalised face regions resized to the required input resolution; frame-level embeddings are mean-pooled within each window.  Audio preprocessing depends on the backbone: Wav2Vec2 processes normalised raw waveforms to produce contextualised speech representations, while MAE-Audio encodes time–frequency spectrogram patches via masked autoencoding. In both cases, latent outputs are mean-pooled per window. Multimodal baselines (e.g., MMA-DFER, HiCMAE) preprocess each modality according to their respective unimodal pipelines and output fused audio-visual embeddings aggregated over the same temporal windows. All pretrained encoders are kept frozen to ensure fair comparison.

\paragraph{Preference Pair Construction:}
Preference pairs are formed between consecutive windows $(x_t, x_{t+1})$ based on their averaged affect values $\{a_t\}$. To reduce noise from minor annotation fluctuations, we retain only pairs whose relative affect change exceeds a threshold $\tau$ (10\% or 20\%).  For each valid pair, we include both directional orders  $(x_t, x_{t+1})$ and $(x_{t+1}, x_t)$ with complementary labels to increase supervision and maintain class balance. Larger thresholds and longer temporal windows naturally reduce the number of valid pairs, yielding between 1,326 and 6,732 training samples depending on the configuration. The resulting difference representations are used to train the preference learner described in Section~\ref{sec:preference_learner}.

\section{Experiments}
\subsection{Experimental Protocol}

Experiments target directional valence and arousal change prediction under a subject-independent protocol, ensuring that recordings from the same participant never appear in both training and evaluation sets. For SEWA, we perform 15-fold subject-level cross-validation, while for Aff-Wild2 we run 15 independent seeds; results report the mean across folds or runs. Preference pairs are constructed between consecutive temporal windows (3s and 5s) when the relative change in Valence or Arousal exceeds a margin (10\% or 20\%). Both directional orders are included to maintain class balance. Predictions operate on window-level representations using embedding differences between consecutive segments. Classification uses a two-layer MLP with hidden sizes $\left(\min(d/2, 250), \min(d/4, 125)\right)$, where $d$ is the input dimensionality. Training uses Adam (max 25 iterations), $\ell_2$ regularisation $\alpha=1$, tolerance $10^{-3}$, shuffling, and early stopping after three non-improving iterations, typically converging in 8–12 epochs. Optimisation uses binary cross-entropy. Performance is measured using accuracy, with statistical significance assessed via the Wilcoxon signed-rank test ($p < 0.05$).

\subsection{Language Backbone Sensitivity Analysis}

\begin{table}[t]
\centering
\scriptsize
\setlength{\tabcolsep}{3pt}
\renewcommand{\arraystretch}{1.05}
\caption{\textbf{Aff-Wild2 accuracy of the LaScA framework across modalities, window sizes, thresholds and backbones. }``Features'' and ``Sentence Transformers'' denote feature-only and text-only models, respectively. The \textbf{(F)} suffix indicates multimodal fusion of text embeddings and features. \textbf{Bold} values represent the highest accuracy; \underline{underlined} values are statistically on par with the best-performing models.}
\label{tab:aff-wild2-results}
\label{tab:affwild2_all}
\begin{tabular}{l l cc cc cc cc}
\toprule
 &  & \multicolumn{4}{c}{Arousal} & \multicolumn{4}{c}{Valence} \\
\cmidrule(lr){3-6} \cmidrule(lr){7-10}
Modality & Model
& \multicolumn{2}{c}{3s} 
& \multicolumn{2}{c}{5s}
& \multicolumn{2}{c}{3s}
& \multicolumn{2}{c}{5s} \\
\cmidrule(lr){3-4}\cmidrule(lr){5-6}
\cmidrule(lr){7-8}\cmidrule(lr){9-10}
 &  & 10\% & 20\% & 10\% & 20\% & 10\% & 20\% & 10\% & 20\% \\
\midrule

\multirow{11}{*}{Visual}
& Features &\underline{0.65}& 0.59 & \underline{0.69}& 0.61 & \textbf{0.60}& 0.60 & \underline{0.67}& 0.63 \\
& MPNet &0.56 & 0.51 & 0.60 & 0.58 & 0.54 & 0.56 & 0.58 & 0.66 \\
& QAMPNet &0.56 & 0.51 & 0.60 & 0.60 & 0.56 & 0.56 & 0.61 & 0.67 \\
& DRoBERTa &0.58 & 0.50 & 0.62 & 0.64 & 0.56 & 0.53 & 0.59 & 0.65 \\
& MiniLM &0.57 & 0.50 & 0.61 & 0.61 & 0.55 & 0.55 & 0.59 & 0.67 \\
& DBERT &0.58 & 0.54 & 0.63 & 0.64 & 0.55 & 0.56 & 0.59 & 0.65 \\
& MPNet(F) &\textbf{0.66}& \textbf{0.65}& \underline{0.69}& \underline{0.74}& \underline{0.59}& \textbf{0.62}& \textbf{0.68}& \textbf{0.74}\\
& QAMPNet(F) &\underline{0.65}& 0.58 & 0.68 & 0.70 & \underline{0.59}& 0.58 & \underline{0.67}& 0.71 \\
& DRoBERTa(F) &\underline{0.65}& \underline{0.64}& \textbf{0.70}& 0.73 & \underline{0.59}& \underline{0.61}& \underline{0.67}& \textbf{0.74}
\\
& MiniLM(F) &\underline{0.65}& \underline{0.64}& \underline{0.69}& \textbf{0.75}& \underline{0.59}& \underline{0.61}& \underline{0.67}& \underline{0.73}\\
& DBERT(F) &\underline{0.65}& 0.62 & \textbf{0.70}& 0.73 & \underline{0.59}& \underline{0.61}& \textbf{0.68}& \underline{0.73}\\

\midrule

\multirow{11}{*}{Audio}
& Features &0.54 & 0.54 & 0.55 & 0.55 & 0.52 & 0.52 & 0.54 & 0.52 \\
& MPNet &0.52 & 0.57 & 0.55 & 0.60 & 0.51 & 0.54 & 0.53 & 0.53 \\
& QAMPNet &0.52 & 0.58 & 0.55 & 0.60 & 0.52 & \underline{0.55}& 0.52 & 0.54 \\
& DRoBERTa &0.52 & 0.57 & 0.55 & 0.58 & 0.51 & \underline{0.55}& 0.52 & 0.53 \\
& MiniLM &0.52 & 0.58 & 0.55 & 0.58 & 0.51 & 0.54 & 0.53 & 0.54 \\
& DBERT &0.52 & 0.57 & 0.55 & 0.59 & 0.51 & 0.54 & 0.52 & 0.52 \\
& MPNet(F) &\textbf{0.62} & \underline{0.67} & \underline{0.63}& \underline{0.72}& \textbf{0.54}& \textbf{0.56}& \textbf{0.58}& \underline{0.58}\\
& QAMPNet(F) &\textbf{0.62} & \underline{0.67} & \textbf{0.64}& \textbf{0.73}& \textbf{0.54}& \textbf{0.56}& \textbf{0.58}& \underline{0.58}\\
& DRoBERTa(F) &\textbf{0.62 }& \textbf{0.68} & \textbf{0.64}& \underline{0.72}& \textbf{0.54}& \textbf{0.56}& \underline{0.57}& \underline{0.58}\\
& MiniLM(F) &\textbf{0.62} & 0.66 & 0.62 & \underline{0.72}& \textbf{0.54}& \underline{0.55}& \underline{0.57}& \underline{0.58}\\
& DBERT(F) &\textbf{0.62} & \textbf{0.68} & \textbf{0.64}& \underline{0.72}& \textbf{0.54}& \textbf{0.56}& \textbf{0.58}& \textbf{0.59}\\

\midrule

\multirow{11}{*}{Multimodal}
& Features &0.54 & 0.54 & 0.55 & 0.55 & 0.52 & 0.52 & 0.54 & 0.52 \\
& MPNet &0.56 & 0.52 & 0.61 & 0.60 & 0.56 & 0.56 & 0.55 & 0.62 \\
& QAMPNet &0.57 & 0.55 & 0.62 & 0.65 & 0.56 & 0.57 & \underline{0.61}& \textbf{0.67}\\
& DRoBERTa &0.57 & 0.50 & 0.61 & 0.58 & 0.55 & 0.53 & 0.56 & 0.61 \\
& MiniLM &0.58 & 0.52 & 0.60 & 0.55 & 0.56 & 0.55 & 0.56 & 0.62 \\
& DBERT &0.59 & 0.54 & 0.63 & 0.64 & 0.54 & 0.56 & 0.58 & \underline{0.65}\\
& MPNet(F) &\textbf{0.65}& \underline{0.69}& \underline{0.67}& \underline{0.74}& \underline{0.58}& \underline{0.60}& \underline{0.61}& 0.61 \\
& QAMPNet(F) &\underline{0.64}& 0.67 & \textbf{0.68}& \underline{0.74}& \underline{0.58}& \textbf{0.61}& \textbf{0.62}& \underline{0.66}\\
& DRoBERTa(F) &\textbf{0.65}& \textbf{0.70}& \underline{0.67}& \underline{0.74}& \textbf{0.59}& \underline{0.60}& \underline{0.61}& 0.63 \\
& MiniLM(F) &\textbf{0.65}& \underline{0.69}& \underline{0.67}& 0.72 & \underline{0.58}& 0.57 & \underline{0.61}& 0.61 \\
& DBERT(F) &\textbf{0.65}& \underline{0.69}& \underline{0.67}& \textbf{0.75}& \textbf{0.59}& \underline{0.60}& \underline{0.61}& 0.61 \\

\bottomrule
\end{tabular}
\end{table}

\begin{table}[t]
\centering
\scriptsize
\setlength{\tabcolsep}{3pt}
\renewcommand{\arraystretch}{1.05}
\caption{\textbf{SEWA DB accuracy of the LaScA framework across modalities, window sizes, thresholds and backbones.}``Features'' and ``Sentence Transformers'' denote feature-only and text-only models, respectively. The \textbf{(F)} suffix indicates multimodal fusion of text embeddings and features. \textbf{Bold} values represent the highest accuracy; \underline{underlined} values are statistically on par with the best-performing models.}
\label{tab:sewa_all}
\begin{tabular}{l l cc cc cc cc}
\toprule
 &  & \multicolumn{4}{c}{Arousal} & \multicolumn{4}{c}{Valence} \\
\cmidrule(lr){3-6} \cmidrule(lr){7-10}
Modality & Model
& \multicolumn{2}{c}{3s} 
& \multicolumn{2}{c}{5s}
& \multicolumn{2}{c}{3s}
& \multicolumn{2}{c}{5s} \\
\cmidrule(lr){3-4}\cmidrule(lr){5-6}
\cmidrule(lr){7-8}\cmidrule(lr){9-10}
 &  & 10\% & 20\% & 10\% & 20\% & 10\% & 20\% & 10\% & 20\% \\
\midrule

\multirow{11}{*}{Visual}
& Features &0.51 & 0.50 & 0.50 & 0.50 & 0.52 & 0.50 & 0.50 & 0.50 \\
& MPNet &0.54 & 0.56 & 0.53 & 0.68 & 0.57 & 0.63 & 0.61 & 0.74 \\
& QAMPNet &0.58 & 0.57 & 0.55 & 0.67 & 0.58 & 0.68 & 0.68 & 0.70 \\
& DRoBERTa &0.55 & 0.62 & 0.53 & 0.61 & 0.60 & 0.69 & 0.67 & 0.69 \\
& MiniLM &0.57 & 0.50 & 0.52 & 0.50 & 0.58 & 0.56 & 0.64 & 0.53 \\
& DBERT &0.55 & 0.62 & 0.51 & 0.56 & 0.60 & 0.68 & 0.63 & 0.71 \\
& MPNet(F) &\underline{0.64}& 0.61 & \textbf{0.65}& 0.66 & \underline{0.69}& 0.72 & \underline{0.72}& 0.76 \\
& QAMPNet(F) &\textbf{0.65}& 0.61 & \underline{0.64}& 0.62 & 0.63 & 0.71 & \underline{0.72}& 0.75 \\
& DRoBERTa(F) &\underline{0.64}& 0.61 & \underline{0.64}& 0.65 & \textbf{0.70}& 0.73 & \textbf{0.73}& 0.78 \\
& MiniLM(F) &\textbf{0.65}& \textbf{0.69}& \underline{0.64}& \textbf{0.70}& \textbf{0.70}& \textbf{0.78}& \underline{0.72}& \textbf{0.83}\\
& DBERT(F) &\textbf{0.65}& 0.62 & \textbf{0.65}& 0.67 & \underline{0.69}& 0.73 & \underline{0.72}& 0.76 \\

\midrule

\multirow{11}{*}{Audio}
& Features &0.50 & 0.50 & 0.50 & 0.50 & 0.50 & 0.50 & 0.50 & 0.50 \\
& MPNet &0.49 & 0.58 & \underline{0.55}& 0.67 & 0.53 & 0.54 & 0.57 & 0.54 \\
& QAMPNet &0.51 & 0.56 & \underline{0.55}& 0.69 & 0.50 & 0.51 & 0.56 & 0.57 \\
& DRoBERTa &0.51 & 0.58 & \textbf{0.56}& 0.61 & 0.50 & \textbf{0.58}& 0.57 & 0.59 \\
& MiniLM &0.53 & 0.50 & \underline{0.55}& 0.50 & 0.50 & 0.47 & 0.54 & 0.52 \\
& DBERT &0.53 & 0.57 & \underline{0.55}& 0.67 & 0.52 & 0.56 & \underline{0.58}& 0.60 \\
& MPNet(F) &\underline{0.58}& \textbf{0.60}& \underline{0.55}& \underline{0.71}& \textbf{0.57}& 0.56 & \underline{0.58}& \textbf{0.63}\\
& QAMPNet(F) &0.53 & \underline{0.59}& \underline{0.55}& \textbf{0.72}& 0.52 & 0.55 & 0.57 & 0.59 \\
& DRoBERTa(F) &\underline{0.58}& \underline{0.59}& \textbf{0.56}& 0.70 & \underline{0.56}& \underline{0.57}& \textbf{0.59}& \underline{0.62}\\
& MiniLM(F) &\underline{0.58}& \underline{0.59}& \textbf{0.56}& 0.68 & 0.54 & \underline{0.57}& 0.56 & 0.60 \\
& DBERT(F) &\textbf{0.59}& 0.54 & \underline{0.55}& \underline{0.71}& \textbf{0.57}& 0.55 & \textbf{0.59}& \underline{0.62}\\

\midrule

\multirow{11}{*}{Multimodal}
& Features &0.50 & 0.50 & 0.50 & 0.50 & 0.50 & 0.50 & 0.50 & 0.50 \\
& MPNet &0.55 & 0.55 & 0.55 & 0.64 & 0.56 & 0.63 & 0.58 & 0.65 \\
& QAMPNet &0.60 & 0.59 & 0.57 & 0.70 & 0.56 & 0.69 & 0.71 & 0.72 \\
& DRoBERTa &0.54 & 0.53 & 0.51 & 0.59 & 0.56 & 0.64 & 0.63 & 0.68 \\
& MiniLM &0.56 & 0.50 & 0.55 & 0.50 & 0.56 & 0.59 & 0.65 & 0.55 \\
& DBERT &0.53 & 0.62 & 0.51 & 0.58 & 0.58 & 0.66 & 0.67 & 0.73 \\
& MPNet(F) &0.65 & \textbf{0.70}& \underline{0.62}& 0.70 & \textbf{0.71}& \underline{0.78}& \textbf{0.75}& \textbf{0.81}\\
& QAMPNet(F) &0.61 & 0.64 & \underline{0.62}& \underline{0.72}& 0.63 & 0.72 & 0.67& 0.79 \\
& DRoBERTa(F) &\underline{0.66}& 0.66 & 0.59 & \textbf{0.73}& \underline{0.70}& 0.77 & 0.69 & \textbf{0.81}\\
& MiniLM(F) &\underline{0.66}& 0.58 & \textbf{0.63}& 0.53 & \underline{0.69}& 0.67 & 0.71 & 0.59 \\
& DBERT(F) &\textbf{0.67}& \underline{0.68}& 0.61 & 0.68 & \underline{0.69}& \textbf{0.79}& 0.71 & \textbf{0.81}\\

\bottomrule
\end{tabular}
\end{table}

Tables~\ref{tab:affwild2_all} and \ref{tab:sewa_all} report the performance of the LaScA framework on \textit{Aff-Wild2} and \textit{SEWA DB} across modalities, temporal windows (3s/5s), thresholds (10\%/20\%), and backbone architectures, comparing feature-only baselines, text-only Sentence Transformers, and fused variants (F). On Aff-Wild2, multimodal fusion consistently improves performance across visual, audio, and multimodal settings, independent of backbone or threshold. For the \textbf{visual modality}, fusion raises arousal up to \textbf{0.75} (5s/20\%) and valence up to \textbf{0.74}, indicating that linguistic context helps disambiguate subtle facial expressions. Larger gains with 5s windows highlight the benefit of temporally extended semantic context. For \textbf{audio}, unimodal performance is moderate, but fusion increases arousal to \textbf{0.73}, showing that textual embeddings provide complementary cues that stabilise predictions. In the \textbf{multimodal setting}, fusion improves peak accuracy and reduces variance across thresholds, yielding more stable affect representations.

On SEWA DB, fusion has an even stronger impact. Unimodal baselines, particularly visual, often operate near chance, while fused models reach valence up to \textbf{0.83}, demonstrating the need for contextual modelling in conversational interactions. Textual embeddings compensate for subtle facial cues and ambiguous prosody, leading to consistently high multimodal performance across backbones. This convergence suggests that, once semantic alignment is introduced, the integration strategy dominates over backbone choice. Across both datasets, three patterns emerge: (i) fusion provides larger relative gains on SEWA DB, highlighting effectiveness in unconstrained, conversational scenarios; (ii) 5-second windows generally outperform 3-second windows, emphasising the value of extended temporal context; (iii) improvements are stronger for arousal, suggesting semantic grounding particularly enhances affective intensity modelling. Overall, LaScA leverages semantic alignment to produce stable, context-aware affect representations, yielding consistent improvements across datasets, modalities, and architectures.

\subsection{Comparison with State-of-the-Art}
Tables~\ref{tab:sota-comparison-affeild} and \ref{sota-sewa} compare LaScA (MPNet backbone) with recent state-of-the-art approaches on \textit{Aff-Wild2} and \textit{SEWA DB}, reporting results across modalities, temporal windows, and thresholds. On Aff-Wild2, LaScA performs consistently at or above strong baselines across all modalities. In the \textbf{visual modality}, it matches or slightly surpasses SwinFace and MAE-Face, reaching \textbf{0.74} arousal and valence accuracy (5s/20\%), while maintaining statistical parity in most other configurations. For \textbf{audio}, LaScA is competitive with Wav2Vec2 and MAE-Audio, achieving \textbf{0.72} arousal at 5s/20\% despite relying only on pre-extracted features rather than end-to-end self-supervised audio. In the \textbf{multimodal setting}, LaScA performs on par with HiCMAE and MMA-DFER, delivering balanced gains across both affect dimensions and temporal windows.

On SEWA DB, comparison is restricted to \textbf{visual models} due to the lack of raw audio. LaScA achieves the highest valence accuracy (\textbf{0.83}, 5s/20\%) and remains statistically comparable for arousal across configurations. While some transformers achieve higher peak arousal in short windows, LaScA provides more consistent valence improvements, suggesting semantic grounding is especially effective for disambiguating affect polarity in conversational scenarios. Across both datasets, LaScA demonstrates state-of-the-art-level performance without modality- or backbone-specific tuning. Its gains arise from a unified semantic alignment mechanism, which is particularly beneficial for valence prediction and longer temporal windows, highlighting the importance of contextual and linguistic cues in modelling affective dynamics.

\begin{table}[t]
\centering
\scriptsize
\setlength{\tabcolsep}{3pt}
\renewcommand{\arraystretch}{1.05}
\caption{Quantitative comparison of our method with state-of-the-art approaches for Aff-Wild2 prediction. \textbf{Bold} values indicate the best-performing model, while \underline{underlined} values indicate models performing statistically on par.}
\label{tab:sota-comparison-affeild}
\begin{tabular}{l l cc cc cc cc}
\toprule
 &  & \multicolumn{4}{c}{Arousal} & \multicolumn{4}{c}{Valence} \\
\cmidrule(lr){3-6} \cmidrule(lr){7-10}
Modality & Model
& \multicolumn{2}{c}{3s} 
& \multicolumn{2}{c}{5s}
& \multicolumn{2}{c}{3s}
& \multicolumn{2}{c}{5s} \\
\cmidrule(lr){3-4}\cmidrule(lr){5-6}
\cmidrule(lr){7-8}\cmidrule(lr){9-10}
 &  & 10\% & 20\% & 10\% & 20\% & 10\% & 20\% & 10\% & 20\% \\
\midrule

\multirow{4}{*}{Visual}
& VGGFace2 & 0.64 & 0.63 & 0.66 & 0.71 & 0.57 & 0.60 & 0.65 & 0.72 \\
& SwinFace & \textbf{0.67} & \textbf{0.67} & \underline{0.70} & \textbf{0.74} & \textbf{0.61} & \textbf{0.63} & \textbf{0.69} & \underline{0.73} \\
& MAE-Face & 0.65 & 0.64 & \textbf{0.71} & 0.72 & 0.58 & \underline{0.61} & \underline{0.67} & 0.71 \\
& LaScA (ours)    & \underline{0.66} & 0.65 & \underline{0.69} & \textbf{0.74} & {0.59} & \underline{0.62} & \underline{0.68} & \textbf{0.74} \\
\midrule

\multirow{3}{*}{Audio}
& Wav2Vec2  & \textbf{0.64} & \underline{0.66} & \underline{0.62} & \underline{0.71} &\textbf{ 0.55} & \textbf{0.57} & \textbf{0.58} & \textbf{0.60} \\
& MAE-Audio & 0.61 & 0.65 & 0.61 & 0.69 & 0.53 & 0.55 & \underline{0.57} & \textbf{0.60} \\
& LaScA (ours) &{0.62} & \textbf{0.67} & \textbf{0.63}& \textbf{0.72}& \underline{0.54}& \underline{0.56}& \textbf{0.58}& {0.58}\\
\midrule

\multirow{3}{*}{Multimodal}

& HiCMAE     & \underline{0.64} & \underline{0.68} & \textbf{0.70} & \textbf{0.75} & \textbf{0.60} & \textbf{0.61} & \textbf{0.61} & \textbf{0.63} \\
& MMA-DFER   & 0.62 & \textbf{0.69} & 0.68 & \textbf{0.75} & \textbf{0.60} & \underline{0.60} & \underline{0.60} & \textbf{0.63} \\
& LaScA (ours) &\textbf{0.65}& \textbf{0.69}& {0.67}& \underline{0.74}& \underline{0.58}& \underline{0.60}& \textbf{0.61}& 0.61 \\
\bottomrule
\end{tabular}
\end{table}

\begin{table}[t]
\centering
\scriptsize
\setlength{\tabcolsep}{3pt}
\renewcommand{\arraystretch}{1.05}
\caption{Quantitative comparison of our method with state-of-the-art approaches for SEWA DB prediction. \textbf{Bold} values indicate the best-performing model, while \underline{underlined} values indicate models performing statistically on par.}
\label{sota-sewa}
\begin{tabular}{l l cc cc cc cc}
\toprule
 &  & \multicolumn{4}{c}{Arousal} & \multicolumn{4}{c}{Valence} \\
\cmidrule(lr){3-6} \cmidrule(lr){7-10}
Modality & Model
& \multicolumn{2}{c}{3s} 
& \multicolumn{2}{c}{5s}
& \multicolumn{2}{c}{3s}
& \multicolumn{2}{c}{5s} \\
\cmidrule(lr){3-4}\cmidrule(lr){5-6}
\cmidrule(lr){7-8}\cmidrule(lr){9-10}
 &  & 10\% & 20\% & 10\% & 20\% & 10\% & 20\% & 10\% & 20\% \\
\midrule

\multirow{4}{*}{Visual}
& VGGFace2 & {0.62} & {0.66} & {0.61} & {0.67} & {0.66} & {0.74} & {0.69} & {0.79} \\
& SwinFace & \textbf{0.67} & \textbf{0.70} & \underline{0.65} & \textbf{0.71} & \textbf{0.72} & \textbf{0.79} & {\textbf{0.73}} & \underline{0.82} \\
& MAE-Face & {0.64} & {0.68} & \textbf{0.66} & \underline{0.70} & {0.69} & {0.77} & \underline{0.71} & {0.81} \\
& LaScA (ours)   & {0.65} & \underline{0.69} & {0.64} & \underline{0.70} & {0.70} & \underline{0.78} & \underline{0.72} & {\textbf{0.83}} \\


\bottomrule
\end{tabular}
\end{table}

\subsection{The Impact of Lexicon}

Table~\ref{tab:lexicon_comparison} reports an ablation on Aff-Wild2 assessing the impact of the lexicon construction strategy in LaScA. We compare a feature-name lexicon against our LLM-generated lexicon under identical multimodal settings, window sizes (3s/5s), and thresholds (10\%/20\%). The LLM-based lexicon consistently matches or improves performance across all configurations. For \textbf{arousal}, gains of about 2\% are observed in most settings. For \textbf{valence}, improvements are smaller but consistent (1–2\%), with the LLM lexicon achieving the best results in all cases. The stronger and more stable gains for arousal suggest that richer lexical grounding particularly benefits modelling affective intensity, where contextual cues influence perceived activation. Overall, these results indicate that LaScA’s performance improvements stem not only from multimodal fusion but also from the quality of the semantic lexicon used.

\begin{table}[t]
\centering
\scriptsize
\setlength{\tabcolsep}{3pt}
\renewcommand{\arraystretch}{1.05}
\caption{Quantitative comparison of feature-based and LLM-based Lexicon approach for Aff-Wild2. Results are reported across Arousal and Valence dimensions for different window sizes (3s, 5s) and selection thresholds (10\%, 20\%). \textbf{Bold} values indicate the best-performing model for each configuration.}
\label{tab:lexicon_comparison}
\begin{tabular}{l l cc cc cc cc}
\toprule
 &  & \multicolumn{4}{c}{Arousal} & \multicolumn{4}{c}{Valence} \\
\cmidrule(lr){3-6} \cmidrule(lr){7-10}
Multimodal & LaScA
& \multicolumn{2}{c}{3s} 
& \multicolumn{2}{c}{5s}
& \multicolumn{2}{c}{3s}
& \multicolumn{2}{c}{5s} \\
\cmidrule(lr){3-4}\cmidrule(lr){5-6}
\cmidrule(lr){7-8}\cmidrule(lr){9-10}
 &  & 10\% & 20\% & 10\% & 20\% & 10\% & 20\% & 10\% & 20\% \\
\midrule

\multirow{2}{*}{MPNet(F)}
& Feature  &0.63 & 0.67 & 0.65 &\textbf{ 0.74} & 0.56 & 0.59 & 0.60 & \textbf{0.61} \\
& LLM (ours) &  \textbf{0.65}& \textbf{0.69}& \textbf{0.67}& \textbf{0.74}& \textbf{0.58}& \textbf{0.60}& \textbf{0.61}& \textbf{0.61} \\
\midrule
\multirow{2}{*}{DRoBERTa(F)}
& Feature  &0.64 & 0.68 & 0.66 & \textbf{0.74} & 0.56 & 0.59 & 0.60 & 0.61 \\
& LLM (ours) &\textbf{0.65}& \textbf{0.70}& \textbf{0.67}& \textbf{0.74}& \textbf{0.59}& \textbf{0.60}& \textbf{0.61}& \textbf{0.63} \\
\midrule
\multirow{2}{*}{MiniLM(F)}
& Feature  &0.64 & 0.67 & 0.65 & \textbf{0.72 }& 0.56 & \textbf{0.57} & 0.60 & 0.60 \\

& LLM (ours) &\textbf{0.65}& \textbf{0.69}& \textbf{0.67}& \textbf{0.72 }& \textbf{0.58}& \textbf{0.57} & \textbf{0.61}& \textbf{0.61} \\

\bottomrule
\end{tabular}
\end{table}

\subsection{Computational Efficiency}
A key design objective of LaScA is to transfer semantic structure from large language models into lightweight affect predictors without introducing the computational overhead of end-to-end deep architectures. LaScA operates on frozen sentence-transformer encoders and trains only a small preference-learning MLP. Depending on the modality and representation dimensionality, the number of trainable parameters ranges from 129 to 230k, keeping the learning component extremely compact. The underlying sentence-transformer backbones remain frozen and range from 22M to 110M parameters, acting purely as semantic feature extractors. Despite this, inference remains efficient. On a workstation equipped with an 11th Gen Intel Core i5 CPU, 16 GB RAM, and an NVIDIA RTX 3060 laptop GPU, end-to-end inference time ranges between 80–140 ms per sample, depending on the encoder used. These results demonstrate that LaScA enables language-conditioned affect modelling while keeping the trainable component lightweight and computationally efficient, effectively transferring semantic structure from large models to compact predictors.
\section{Discussion}

While LaScA demonstrates consistent improvements in affect prediction across modalities and datasets, several aspects offer opportunities for further exploration. First, all pretrained textual, visual, and audio encoders are used as frozen backbones rather than adapted to the affect prediction task. This design isolates the effect of language-conditioned representations and ensures a controlled comparison, but future work could investigate selective fine-tuning or lightweight adaptation strategies to further improve performance. Second, the SEWA DB experiments rely on pre-extracted acoustic features, which prevents evaluation of end-to-end raw waveform models in multimodal settings. Integrating raw-audio encoders could enable richer modelling of prosodic dynamics and may further strengthen multimodal affect representations.

Third, the semantic lexicon is generated once and kept fixed throughout training and evaluation. This choice improves stability, reproducibility, and interpretability, but adaptive or multilingual lexicons could extend LaScA to new domains, languages, or cultural contexts. Similarly, while our experiments compare affect-aware and feature-only lexicons, additional perturbation studies such as shuffled feature description mappings or alternative template structures could provide further insight into how semantic grounding contributes to model performance. Fourth, the current salience mechanism selects active features using a deterministic thresholding procedure. Although this approach provides interpretable cue selection, future work could explore alternative feature selection strategies (e.g., learned gating or top-$k$ selection) to better understand the role of salience in language-conditioned representations.

Finally, LaScA models affect change through pairwise ordinal comparisons between consecutive temporal windows. While effective for capturing local affect dynamics, incorporating longer-range temporal modelling such as sequence encoders or temporal attention could enable richer representations of emotional trajectories. Beyond valence and arousal prediction, extending LaScA to discrete emotion categories or higher-dimensional affect representations also remains a promising direction. Overall, these observations highlight several avenues for expanding the flexibility and expressiveness of LaScA while preserving its key advantages in interpretability, efficiency, and semantic grounding.

\section{Conclusions}

In this study we introduced LaScA, a compact and interpretable framework for modelling affective dynamics in-the-wild by leveraging semantic alignment with pretrained language models. Unlike end-to-end deep architectures, LaScA preserves handcrafted visual and acoustic features while converting them into a frozen semantic lexicon, enabling deterministic language-driven priors over valence and arousal. Experiments on \textit{Aff-Wild2} and \textit{SEWA DB} demonstrate that LaScA consistently improves unimodal and multimodal predictions, with larger gains in conversational or less constrained scenarios and longer temporal windows. Overall, LaScA provides a scalable, transparent, and modality-agnostic alternative to conventional affect modelling pipelines.

\section*{Acknowledgments}
This work has been partly supported by the University of Piraeus Research Center.

{
    \small
    \bibliographystyle{ieeenat_fullname}
    \bibliography{main}
}

\clearpage
\setcounter{page}{1}
\maketitlesupplementary

\appendix
\section{LLM Instruction Prompt}\label{app:llm_prompt}
\label{app:prompt}
\begin{tcolorbox}[
title=Instruction Prompt for Offline Semantic Lexicon Construction,
colback=gray!5,
colframe=black!20,
boxrule=0.5pt,
arc=2pt,
left=4pt,
right=4pt,
top=4pt,
bottom=4pt,
]
\begin{lstlisting}
You are an expert in affective computing and multimodal emotion modelling.

Your task is to convert low-level video (facial blendshapes) and audio (acoustic features such as MFCCs) feature names into compact semantic labels that combine:

1) A brief physical cue
2) A brief affect cue

STRICT RULES:
1. Output must be a valid Python dictionary.
2. No explanations outside the dictionary.
3. Each value 3 to 6 words.
4. No full sentences.
5. No punctuation inside values.
6. Format: "<expression cue> <affect cue>"
7. Video: describe visible movement.
8. Audio: describe acoustic change.
9. Compact affect cue.
10. Avoid deterministic language.
11. Keep vocabulary consistent.
12. Optimise for SentenceTransformer embeddings.
13. Prioritise reproducibility.

Example format:
'feature_name': 'feature meaning and affect indication'

Return only the dictionary.
\end{lstlisting}
\end{tcolorbox}

\section{Description Generation}
\subsection{Audio Descriptors}\label{app:audio_desc}
\begin{table}[h]
\centering
\footnotesize
\setlength{\tabcolsep}{4pt}
\renewcommand{\arraystretch}{1.05}
\caption{Affect-aware semantic labels for MFCC acoustic features generated offline.}
\label{app:mfcc_lexicon}
\begin{tabular}{p{0.38\columnwidth} p{0.52\columnwidth}}
\toprule
\textbf{Feature Name} & \textbf{Semantic Label} \\
\midrule
mfcc\_0  & high arousal energy \\
mfcc\_1  & dominant low tone \\
mfcc\_2  & neutral stability \\
mfcc\_3  & tense vocal focus \\
mfcc\_4  & clear controlled tone \\
mfcc\_5  & urgent brightness \\
mfcc\_6  & strained timbre \\
mfcc\_7  & assertive pressure \\
mfcc\_8  & excited anxiety \\
mfcc\_9  & tight emphasis \\
mfcc\_10 & breathy nervousness \\
mfcc\_11 & withdrawn low energy \\
mfcc\_12 & piercing tension \\
\bottomrule
\end{tabular}
\end{table}
\newpage
\subsection{Facial Descriptors}\label{app:facail_desc}

\begin{table}[h]
\centering
\scriptsize
\setlength{\tabcolsep}{2.5pt}
\renewcommand{\arraystretch}{1.02}
\caption{Affect-aware semantic lexicon for facial blendshape features.}
\label{app:face_lexicon}
\begin{tabularx}{\columnwidth}{>{\ttfamily}p{0.42\columnwidth} X}
\toprule
\textbf{Feature Name} & \textbf{Semantic Label} \\
\midrule
Face\_browDownLeft & left brow down focused irritation \\
Face\_browDownRight & right brow down skeptical tension \\
Face\_browInnerUp & inner brows up sad vulnerability \\
Face\_browOuterUpLeft & left outer brow up curious surprise \\
Face\_browOuterUpRight & right outer brow up questioning surprise \\
Face\_cheekPuff & cheeks puff held frustration \\
Face\_cheekSquintLeft & left cheek tight restrained smile \\
Face\_cheekSquintRight & right cheek tight restrained smile \\
Face\_eyeBlinkLeft & left blink stress regulation \\
Face\_eyeBlinkRight & right blink stress regulation \\
Face\_eyeLookDownLeft & left gaze down shame reflection \\
Face\_eyeLookDownRight & right gaze down shame reflection \\
Face\_eyeSquintLeft & left eye squint suspicious focus \\
Face\_eyeSquintRight & right eye squint evaluative doubt \\
Face\_eyeWideLeft & left eye wide alarm arousal \\
Face\_eyeWideRight & right eye wide heightened alertness \\
Face\_jawForward & jaw thrust assertive dominance \\
Face\_jawLeft & jaw left uneasy tension \\
Face\_jawOpen & jaw drop shock surprise \\
Face\_jawRight & jaw right uneasy tension \\
Face\_mouthClose & lips closed emotional restraint \\
Face\_mouthPressLeft & left lip press suppressed anger \\
Face\_mouthPressRight & right lip press controlled tension \\
Face\_mouthRollLower & lower lip roll inhibited emotion \\
Face\_mouthRollUpper & upper lip roll inhibited emotion \\
Face\_mouthFrownLeft & left corner down sad disapproval \\
Face\_mouthFrownRight & right corner down sad disapproval \\
Face\_mouthLowerDownLeft & left lower lip down vulnerable sadness \\
Face\_mouthLowerDownRight & right lower lip down vulnerable sadness \\
Face\_mouthSmileLeft & left smile smirk positivity \\
Face\_mouthSmileRight & right smile genuine positivity \\
Face\_mouthDimpleLeft & left dimple warm engagement \\
Face\_mouthDimpleRight & right dimple warm engagement \\
Face\_mouthStretchLeft & left mouth stretch awkward tension \\
Face\_mouthStretchRight & right mouth stretch awkward tension \\
Face\_mouthFunnel & mouth funnel uncertain anticipation \\
Face\_mouthPucker & lip pucker affection hesitation \\
Face\_mouthShrugLower & lower lip shrug doubt uncertainty \\
Face\_mouthShrugUpper & upper lip shrug skeptical hesitation \\
Face\_mouthUpperUpLeft & left upper lip up disgust contempt \\
Face\_mouthUpperUpRight & right upper lip up disgust contempt \\
Face\_noseSneerLeft & left nose sneer strong disgust \\
Face\_noseSneerRight & right nose sneer strong disgust \\
\bottomrule
\end{tabularx}
\end{table}

\subsection{Description Template}
\label{app:description_template}

For each temporal window, textual descriptions are constructed deterministically from the activated handcrafted features using predefined semantic mappings and structured templates.

\paragraph{Unimodal Facial Template.}
Facial blendshape features are first mapped to compact affect-aware semantic labels using the fixed lexicon described in Section~3.1. Given the subset of active facial features for a window, their corresponding semantic labels are concatenated into a structured textual representation via the \texttt{FacialLLMDescriptionGenerator}. The generator produces a consistent phrase ordering to ensure stable downstream embeddings. The resulting facial prompt takes the form:

\begin{equation}
T^{\text{face}}_t = \texttt{concat}(\ell_{i_1}, \ell_{i_2}, \dots, \ell_{i_k}) \ \texttt{<|endoftext|>}
\end{equation}

where $\ell_{i_j}$ denotes the semantic label associated with the $j$-th activated facial feature in window $t$.

\paragraph{Unimodal Audio Template.}
Audio features (MFCC-derived descriptors) are processed analogously using the \texttt{AudioLLMDescriptionGenerator}. Activated acoustic features are converted into their corresponding semantic labels and concatenated into a structured textual prompt:

\begin{equation}
T^{\text{audio}}_t = \texttt{concat}(a_{i_1}, a_{i_2}, \dots, a_{i_m}) \ \texttt{<|endoftext|>}
\end{equation}

where $a_{i_j}$ denotes the semantic label of the $j$-th active acoustic feature.

\paragraph{Multimodal Template Construction.}
The multimodal description is formed by concatenating the facial and audio prompts while preserving ordering consistency. The intermediate end-of-text markers are removed to avoid fragmentation, and a single terminal token is appended:

\begin{equation}
T^{\text{multi}}_t =
\texttt{strip}(T^{\text{face}}_t) \ \texttt{|} \
\texttt{strip}(T^{\text{audio}}_t) \ \texttt{<|endoftext|>}
\end{equation}

This construction ensures: (i) deterministic prompt generation, (ii) stable formatting across samples, (iii) modality-aware separation via the delimiter ``|'', and (iv) compatibility with sentence-transformer tokenization. All templates are generated offline and remain fixed throughout training and evaluation to ensure full reproducibility.

\subsection{Description Examples}

To illustrate the structure of the language-conditioned representations used in \textbf{LaScA}, we report the five most frequent prompts observed in the dataset. Prompts are constructed by combining active facial descriptors with acoustic cues.

\begin{footnotesize}
\begin{enumerate}

\item \texttt{facial: left blink stress regulation, right blink stress regulation, left gaze down shame reflection, right gaze down shame reflection, left eye squint suspicious focus, right eye squint evaluative doubt}\\
\texttt{audio: Acoustic markers indicate high arousal energy.}

\item \texttt{facial: left brow down focused irritation, right brow down skeptical tension, left blink stress regulation, right blink stress regulation, left gaze down shame reflection, right gaze down shame reflection, left eye squint suspicious focus, right eye squint evaluative doubt}\\
\texttt{audio: Acoustic markers indicate high arousal energy.}

\item \texttt{facial: inner brows up sad vulnerability, left outer brow up curious surprise, right outer brow up questioning surprise, left gaze down shame reflection, right gaze down shame reflection}\\
\texttt{audio: Acoustic markers indicate high arousal energy.}

\item \texttt{facial: inner brows up sad vulnerability, left outer brow up curious surprise, right outer brow up questioning surprise}\\
\texttt{audio: Acoustic markers indicate high arousal energy.}

\item \texttt{facial: inner brows up sad vulnerability, left outer brow up curious surprise, right outer brow up questioning surprise, left blink stress regulation, right blink stress regulation, left gaze down shame reflection, right gaze down shame reflection, left eye squint suspicious focus, right eye squint evaluative doubt}\\
\texttt{audio: Acoustic markers indicate high arousal energy.}

\end{enumerate}
\end{footnotesize}

\begin{figure}[t]
\centering
\includegraphics{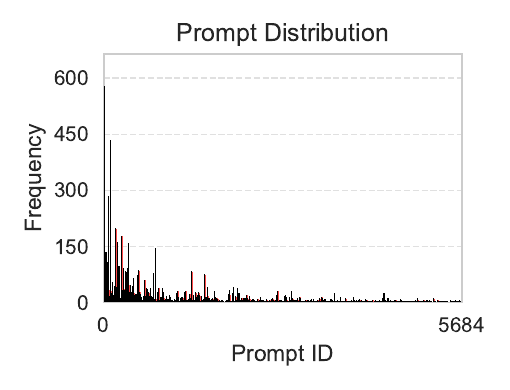}
\caption{
Histogram of unique multimodal prompts
}
\label{fig:prompt_hist}
\end{figure}

\end{document}